\documentclass[journal]{IEEEtran}

\ifCLASSINFOpdf
  \usepackage[pdftex]{graphicx}
\else

\fi

\usepackage{amsmath}

\usepackage{url}

\usepackage{hyperref}
\usepackage{subfigure}
\usepackage{makeidx}  
\usepackage{graphicx}
\usepackage{amsmath}
\usepackage{nccmath}
\usepackage{extarrows}
\usepackage{amsfonts,amsmath}
\usepackage{algorithm2e}
\usepackage{changepage}
\usepackage{float}
\usepackage{multirow}
\usepackage{amssymb}

\usepackage{multirow}
\usepackage{cite}

\begin{document}
%
\title{Real-time Nonrigid Mosaicking of\\ Laparoscopy Images}
%
%
%

\author{Haoyin Zhou,~\IEEEmembership{Member,~IEEE,}
        and Jagadeesan Jayender,~\IEEEmembership{Senior Member,~IEEE}
\thanks{This work was supported by the National Institute of Biomedical Imaging and Bioengineering of the National Institutes of Health through Grant Numbers K99EB027177, R01EB025964 and P41EB015898. Jagadeesan Jayender owns equity in Navigation Sciences, Inc. He is a co-inventor of a navigation device to assist surgeons in tumor excision that is licensed to Navigation Sciences. His interests were reviewed and are managed by BWH and Partners HealthCare in accordance with their conflict of interest policies.

Haoyin Zhou and Jagadeesan Jayender are with the Surgical Planning Laboratory, Department of Radiology, Brigham and Women's Hospital, Harvard Medical School,
MA, 02115, USA. e-mail: zhouhaoyin and jayender@bwh.harvard.edu.}
}

%
%

\markboth{Journal of \LaTeX\ Class Files,~Vol.~14, No.~8, August~2015}%
{Shell \MakeLowercase{\textit{et al.}}: Bare Demo of IEEEtran.cls for IEEE Journals}
%



\maketitle

\begin{abstract}
The ability to extend the field of view of laparoscopy images can help the surgeons to obtain a better understanding of the anatomical context. However, due to tissue deformation, complex camera motion and significant three-dimensional (3D) anatomical surface, image pixels may have non-rigid deformation and traditional mosaicking methods cannot work robustly for laparoscopy images in real-time. To solve this problem, a novel two-dimensional (2D) non-rigid simultaneous localization and mapping (SLAM) system is proposed in this paper, which is able to compensate for the deformation of pixels and perform image mosaicking in real-time. The key algorithm of this 2D non-rigid SLAM system is the expectation maximization and dual quaternion (EMDQ) algorithm, which can generate smooth and dense deformation field from sparse and noisy image feature matches in real-time. An uncertainty-based loop closing method has been proposed to reduce the accumulative errors. To achieve real-time performance, both CPU and GPU parallel computation technologies are used for dense mosaicking of all pixels. Experimental results on \textit{in vivo} and synthetic data demonstrate the feasibility and accuracy of our non-rigid mosaicking method.

\end{abstract}


%
\IEEEpeerreviewmaketitle

\section{Introduction}

\IEEEPARstart{M}{inimally} invasive surgeries (MIS) are beneficial for patients due to less trauma, lower blood loss and faster recovery. MIS usually uses the laparoscope as an intraoperative imaging modality to provide surgical guidance. However, due to the limited field of view (FOV) and complex 6-DoF motion of the laparoscope, it is difficult for the surgeons to relate between the laparoscopic images and the \textit{in vivo} anatomical structures \cite{bergen2014stitching}. For example, for hernia repairs, it is imperative for general surgeons to scan the entire region to identify the size of the mesh that needs to be placed. A small FOV provided by the laparoscope makes it challenging to identify the correct size of the mesh. For lung segmentectomy surgery, the surgeon needs to identify different segments of the lung to isolate the segment with the tumor. A larger FOV obtained by image mosaicking technologies will help plan the sublobar resection to ensure complete tumor removal while preserving as much uninvolved lung parenchyma as possible.

Image mosaicking technologies have their roots in the computer vision field. Most early image mosaicking methods align the images according to the homography-based transformation \cite{szeliski1997creating}, which uses a $3\times 3$ matrix to convert the two-dimensional (2D) pixel coordinates from one image to another. For example, Brown et al. used RANSAC \cite{fischler1981random} and SIFT matches \cite{lowe1999object} to compute the homography matrix between images \cite{brown2007automatic}, and this work has also been used to mosaic laryngoscopic images \cite{iakovidis2013efficient}. Bano et al. proposed to estimates the homography using deep learning \cite{bano2020deep}. Homography assumes that the camera motion mainly comprises of rotational motion and/or the environment is planar, which makes it impracticle to handle image parallax caused by translational camera motion in the three-dimensional (3D) environment. To solve this problem, many image mosaicking works have been proposed to integrate local warp functions \cite{zaragoza2013projective}\cite{li2017quasi}\cite{li2015dual}. For example, Zhang et al. proposed a parallax-tolerant image stitching method that first obtains homography-based alignment and then performs content preserving warping for refinement \cite{zhang2014parallax}. Lee et al. developed a local warping method using multiple homographies and warping residuals \cite{lee2020warping}. Chen et al. proposed to use grid mesh to guide the local warp models \cite{chen2016natural}. Fan et al. proposed to use stereo videos for refining the warps \cite{fan2019stereoscopic}. Although these works have shown promising results, they are mostly designed for static environments and are difficult to handle the deforming \textit{in vivo} environments. In addition, the heavy computational burden to compute the local warps made them too slow for real-time surgical navigation.

Image mosaicking methods have found many clinical applications, such as microscopic and fetoscopic images mosaicking. For microscopic images, the main difficulty is to handle the tissue deformation \cite{kose2017automated}. For example, a recent work by Guo et al. proposed to mosaic laser endomicroscopic images by first estimating the initial rigid transformation from feature matches, and then estimating the local non-rigid warps using an intensity-based similarity metric \cite{gong2020robust}. Vercauteren et al. proposed to compensate for the motion distortion of fibered microscopy images by modeling the relationship between motion and motion distortions \cite{vercauteren2006robust}. These methods are sufficient for mosaicking the microscopic images because the objects at the microscopic scale are almost planar, but are difficult to handle laparoscopy images due to the complex 6-DoF camera motion and significant 3D shapes of the tissues. For fetoscopic images, most existing methods consider the tissues as rigid, and the deformation problem has not been not fully addressed \cite{bano2020deepMiccai}\cite{gaisser2018stable}.

In this paper, we propose a novel 2D non-rigid simultaneous localization and mapping (SLAM) method for laparoscopy images mosaicking in real-time. The concept of non-rigid SLAM was proposed in the DynamicFusion work \cite{newcombe2015dynamicfusion}, and is now an emerging topic in the computer vision field. Unlike the traditional rigid SLAM methods that estimate the 6-DoF rigid motion of the camera \cite{totz2011dense}, non-rigid SLAM estimates the deformation and motion of the environment with respect to the camera, which usually has high degrees of freedom. Our 2D non-rigid SLAM method considers the 2D image mosaic as the environment map, which is similar to the 3D point cloud built by traditional 3D SLAM methods. Non-rigid SLAM has also been introduced in the medical field. For example, Song et al. proposed the MISSLAM method to compensate for tissue deformation and track the camera motion in real-time \cite{song2018mis}. However, most existing non-rigid SLAM methods were designed for 3D reconstruction and used the iterative-closet-points (ICP) algorithm for non-rigid alignment, which cannot align 2D images for mosaicking. To solve this problem, one natural idea is to use the feature matches to build non-rigid correspondences between images. However, mismatches are unavoidable and the non-rigid deformation makes it difficult to remove the mismatches. In addition, dense image mosaicking requires the ability to track every pixel, but feature matches are sparse. To overcome these difficulties, we introduce our previous work called as expectation maximization and dual quaternion (EMDQ) \cite{zhou2020smooth}, which can generate dense and smooth deformation field from the sparse and noisy feature matches in real-time. Using EMDQ as the key algorithm to track the deformation of image pixels, we have also proposed an uncertainty-based loop closing method to reduce the accumulative errors. To the best of our knowledge, this is the first non-rigid SLAM method developed for image mosaicking.

The paper is organized as follows: Section II will give a brief introduction of our EMDQ algorithm. Section III will describe the design of the 2D non-rigid SLAM system, including tracking, loop closing, uncertainty management, smoothing and dense mosaicking. Evaluation results on \textit{in vivo} data are presented in Section IV. A discussion is presented in Section V.

\subsection{Related Works}

The field of view of the laparoscope is small and the ability to generate a larger field of view from input laparoscopic videos can provide better guidance for diagnosis and surgical planning, which is clinically useful for several surgical procedures including those on the bladder \cite{soper2012surface}, retina \cite{richa2012hybrid} and colon \cite{karargyris2010three}. This problem has been studied for decades, and the methods can be roughly classified into 3D and 2D. The 3D mosaicking methods are mostly based on structure from motion (SfM) or simultaneous localization and mapping (SLAM). SfM methods require batch processing, thus making it difficult for real-time applications. The SLAM methods, which originated from the robotics navigation field, have been introduced for surgical navigation to extend the field of view of the laparoscope in real-time. For example, our previous work generated dense 3D model of the tissue surface from stereo laparoscopy videos by using SLAM \cite{zhou2019real}. Mahmoud et al. proposed a monocular dense reconstruction method for tissue surfaces \cite{mahmoud2018live}. Mountney et al. proposed to use EKF-SLAM to build a dense 3D textured model of the MIS environment \cite{mountney2009dynamic}. However, because most SLAM systems are designed for static environment, they cannot handle the deformation of soft tissues. To solve this problem, Mountney et al. proposed to learn a periodic organ motion model for deformation compensation \cite{mountney2010motion}, which cannot handle more general tissue deformation. A recent work by Lamarca et al. was able to track the camera motion in deforming environments, but did not perform mosaicking \cite{lamarca2020defslam}. The 3D mosaic can provide richer information to the surgeons than 2D, but often requires additional camera calibration steps in the clinical workflow. Further for stereo laparoscopes, which are not the standard of care imaging modality, the calibration must be highly precise for the stereo matching algorithms \cite{poggi2019guided}. For some clinical applications, 3D mosaicking is less demanding and 2D mosaicking of the laparoscopic images is sufficient for the clinicians to better understand the \textit{in vivo} environment. For example, Erden et al. proposed to quantify the behavior of soft tissues by mosaicking the microlaparoscopic images \cite{erden2012understanding}.

For both 3D and 2D mosaicking, the majority of methods are designed for rigid environments. For 2D cases, the pixel deformation problem is even more serious, because the translational motion of the camera on 3D structures can also contribute to the non-rigid motion of the pixels. An intuitive example is that when the camera scans the tissue surface, distant areas will have slower motion than closer areas in the image, which is also referred to as parallax and has attracted much attention in the computer vision field. For example, Zaragoza et al. proposed an as-projective-as-possible warp that allows local non-projective deviations \cite{zaragoza2013projective}. In the medical field, many works describing algorithms to compensate for pixel deformation have been reported for the mosaicking of microscopy images\cite{hughes2015high}\cite{rosa2012building}\cite{loewke2010vivo}. We refer to the recent review paper \cite{perperidis2020image} by Perperidis et al. for more details. However, since the tissue surface at the microscopic scale is nearly planar, these methods are designed for planar objects with small deformation, which cannot be applied to the laparoscopy images robustly \cite{bergen2014stitching}.

Non-rigid image mosaicking is closely related to a new concept called non-rigid SLAM \cite{newcombe2015dynamicfusion}\cite{innmann2016volumedeform}, which was developed in recent years with the purpose of simultaneously estimating the deformation and performing mosaicking in real-time. Most non-rigid SLAM methods were designed for 3D reconstruction, which converted the deformation estimation problem to an optimization problem that minimized a cost function consisting of ICP, smoothing and other terms, and ran on powerful GPUs to achieve real-time. However, since ICP cannot be applied to RGB images directly without proper segmentation or edge detection, existing non-rigid SLAM methods cannot be used for 2D image mosaicking. To the best of our knowledge, this paper proposes the first non-rigid SLAM system that is purely based on feature matches for non-rigid image mosaicking. In addition, our method can work in real-time on both CPU and GPU.

\section{Brief Introduction of EMDQ}

\begin{figure} [htp]
\vspace{0.0cm}
\centering
  \includegraphics[width=0.5\textwidth]{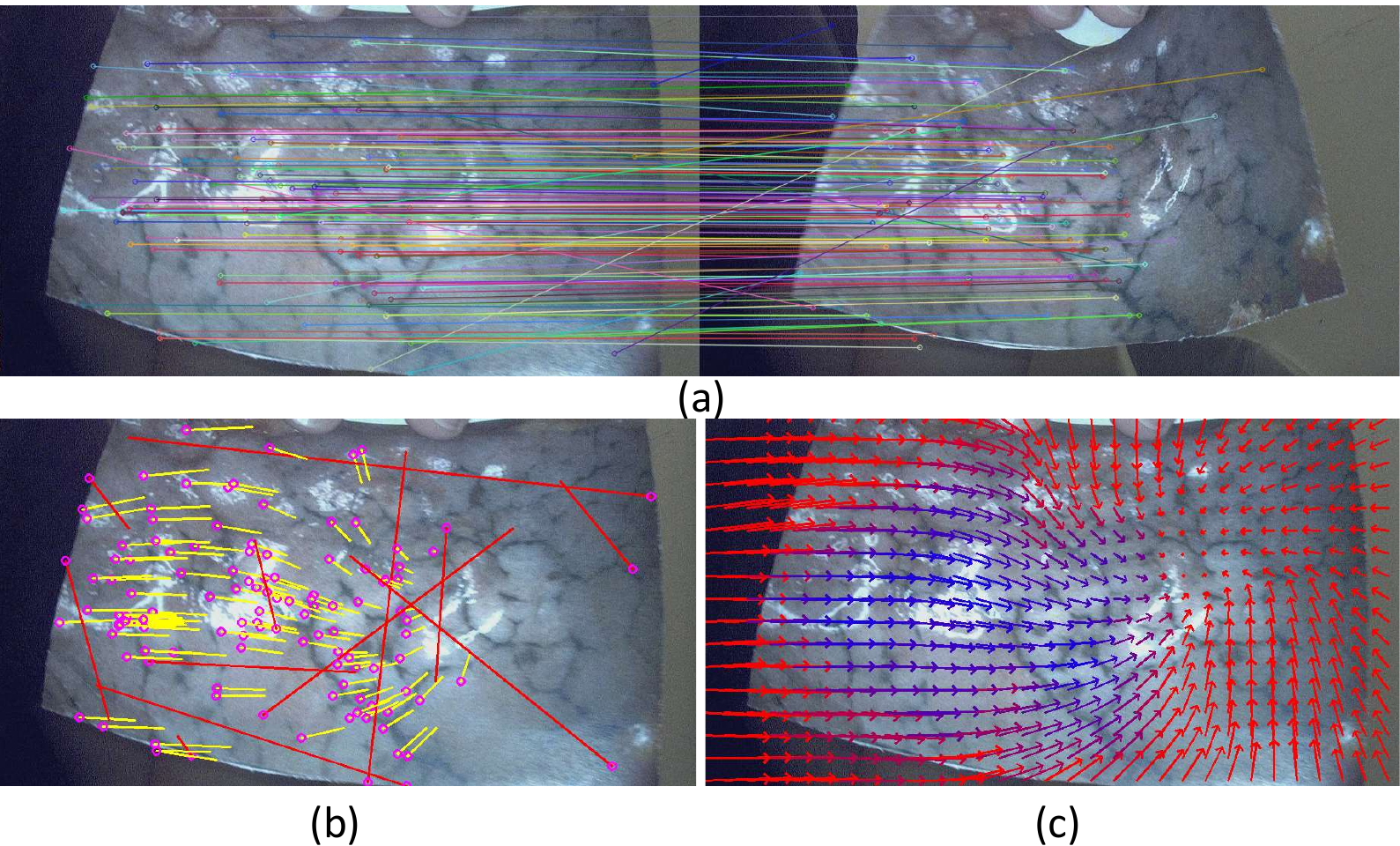}
  \caption{An example of EMDQ with a deformable phantom with lung surface texture. (a) Input SURF matches between two images. (b) The mismatches removal result of EMDQ (yellow: inliers, red: outliers). (c) Smooth and dense deformation field generated by EMDQ in real-time. The color suggests the uncertainties (blue to red: low to high). In this example, the right side of the images has no feaure inliers hence the uncertainty is high. We use a 30 pixels step for clear visualization, but the deformation field and uncertainty can be obtained for every pixel.}
\label{fig_emdqdemo}
\end{figure}

First, we briefly describe the EMDQ algorithm, which is our recent work and the key algorithm in the 2D non-rigid SLAM system proposed in this paper. Details of EMDQ are given in Ref. \cite{zhou2020smooth}. Generally speaking, the EMDQ algorithm is used as a black box in the proposed non-rigid SLAM system. The input of EMDQ is the coordinates of image feature matches and the output is the dense deformation field between images.

\subsection{EMDQ}

The basic idea of the proposed 2D non-rigid SLAM system is to estimate the deformation of image pixels from feature matching results, and then perform mosaicking accordingly. The EMDQ algorithm can remove mismatches from noisy feature matches, and generate smooth and dense deformation field by interpolating among the feature matches in real-time. The deformation field is represented using dual quaternion (DQ) \cite{kavan2008geometric}, which is a useful mathematical tool to generate smooth interpolation among multiple rigid transformations. As an example, Fig. \ref{fig_emdqdemo} shows the motion of pixels from one frame to another that can be obtained by the deformation field generated by EMDQ, which is essential for our non-rigid SLAM system. However, due to the fact that the deformation field is generated by interpolating among sparse feature inliers, two problems need to be further addressed: (1) the feature inliers may not distribute at all areas on the image, hence the deformation field at areas that are distant from feature inliers may be inaccurate, which is often referred to as uncertainty in the registration problem, (2) accumulative errors may exist if the deformation field is tracked from pairs of adjacent video frames. Our 2D non-rigid SLAM consists of multiple algorithms and data management methods to address the above problems.

\subsection{Image Feature Matching Method Selection}

Although EMDQ can generate the deformation field from the results of any image feature matching methods, such as ORB \cite{rublee2011orb} and SURF \cite{bay2006surf}, it is important to select an appropriate feature matching method for the accuracy, speed and robustness of the non-rigid SLAM system. Since EMDQ generates the deformation field by interpolating among the feature matches, the accuracy is higher if all pixels have close feature inliers. Due to this reason, ORB is not appropriate since the ORB feature points mainly distribute at the rich texture areas, which makes it difficult to track low texture areas accurately. Although there exist improved ORB methods \cite{mur2015orb} that are able to detect feature points uniformly on the images and have been widely used in rigid SLAM systems, in practice we found that its percentage of inliers is significantly lower than that of the standard ORB. This is acceptable for rigid SLAM because the rigid motion model can be estimated with a few matches. However for non-rigid SLAM, it may result in low robustness because some image areas may not have feature inliers.

Compared with ORB, SURF is more accurate but much slower, which makes it unpopular for the rigid SLAM systems. A major computational burden of SURF is to build the image pyramid for handling the change of image scale. However, the change of image scale in the laparoscopic image mosaicking task is usually small. In addition, our non-rigid SLAM system tracks the deformation from pairs of adjacent video frames, which have small change of image scale as long as the camera motion is not too fast. Hence, to achieve faster computational speed, we propose to reduce the number of SURF octave layers to one, which avoids the use of image pyramid and significantly reduces the computational burden. In practice we found that SURF with one octave layer works very well. However, given the development of learning-based features \cite{schonberger2017comparative}, we are not implying that SURF is the best choice but it is easy to replace SURF with other features since EMDQ only needs the coordinates of feature matches as the input.

\section{2D Non-rigid SLAM System Design}

\begin{figure*} [htp]
\vspace{0.0cm}
\centering
  \includegraphics[width=1.0\textwidth]{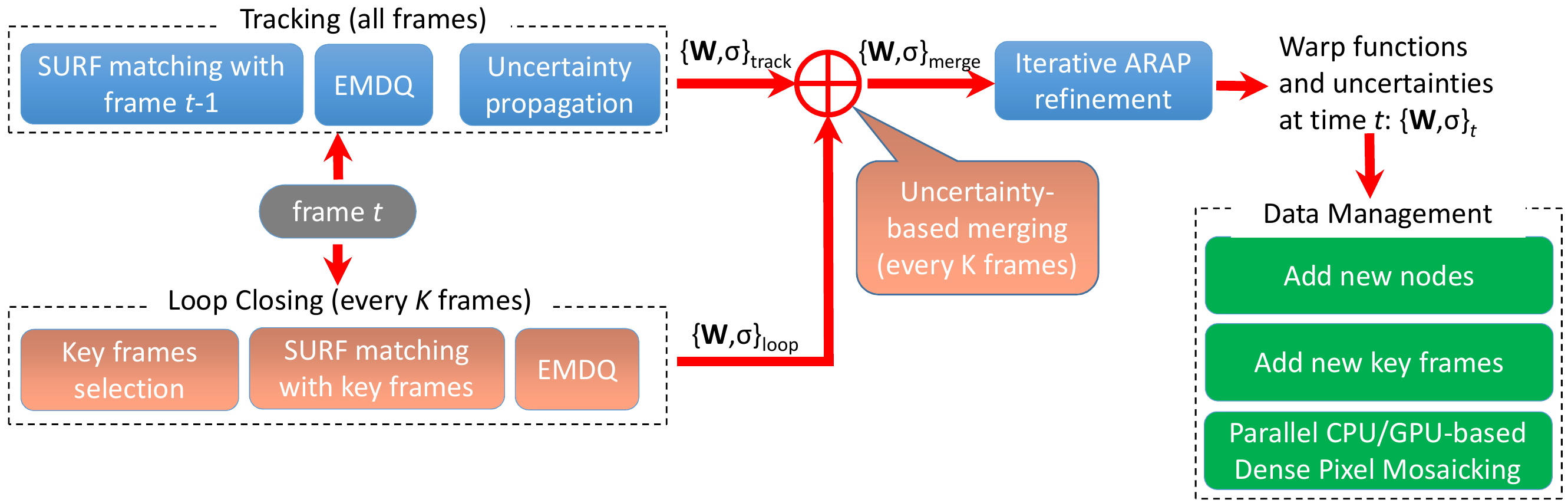}
  \caption{Design of the 2D non-rigid SLAM system. $W$ and $\sigma$ suggest the warp functions and uncertainties of the deformation nodes respectively.}
\label{fig_system}
\end{figure*}

\begin{figure*} [htp]
\vspace{0.0cm}
\centering
  \includegraphics[width=1.0\textwidth]{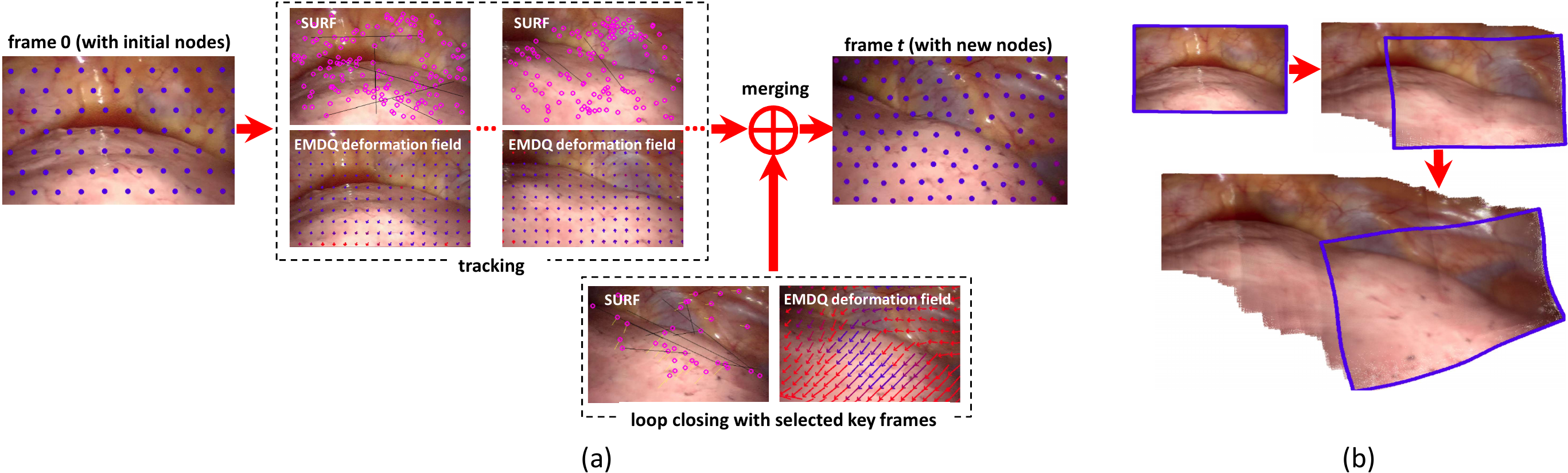}
  \caption{An intuitive example of key steps of our 2D non-rigid SLAM system with a laparoscopic video captured during a lung surgery, which has large deformation due to heartbeat. (a). In both tracking and loop closing, the deformation field are generated by using EMDQ from sparse SURF matches, and the nodes are moved accordingly. The tracking module uses adjacent frames hence more SURF inliers can be found and the deformation is small with low uncertainty. (b) The image mosaicking process. The blue lines are the warped frame edges from time $t$ to 0, which are not straight due to deformation.}.
\label{fig_systemexample}
\end{figure*}

The prerequisite of dense and nonrigid image mosaicking is to recover the deformation of all image pixels. Without loss of generality, this 2D non-rigid SLAM system considers the video frame at time $t=0$ as the reference frame, and estimate the deformation field of frames at $t=1,2,...T$ with respect to frame 0 for mosaicking. Inspired by DynamicFusion \cite{newcombe2015dynamicfusion}, we use sparse control points, or deformation nodes, to represent the deformation field to reduce the computational burden. The nodes are assigned with warp functions and motion of other image pixels are obtained by interpolating among the warp functions of the neighboring deformation nodes. Specifically, the warp function of a node $i = 1,2,...,N$ is represented by a scale factor and a dual quaternion, which is

\begin{equation}
{\bf{x}}_{i,t} = W_{i,t}({\bf{x}}_{i,0}) = s_{i,t} {{\bf{q}}_{i,t}}({\bf{x}}_{i,0})\\ = s_{i,t}({{\bf{R}}_{i,t}}{\bf{x}}_{i,0} + {{\bf{t}}_{i,t}})
\label{eq_warp}
\end{equation}

\noindent where ${\bf{x}}_{i,t} \in \mathbb{R}^2$ and ${\bf{x}}_{i,0} \in \mathbb{R}^2$ are image coordinates of node $i$ at time $t$ and $0$ respectively (in pixels). $W_{i,t}(\cdot)$ represent the warp function of node $i$ at time $t$, and $s_{i,t} \in \mathbb{R}$ and ${{\bf{q}}_{i,t}}\in \mathbb{R}^4$ \footnote{Standard DQ is a 8-dimensional vector. For the image mosaicking task, we consider the image is on the $z=0$ plane and 4 dimensions are always zero, which are removed to reduce the computational burden.} are the related scale factor and dual quaternion respectively. ${{\bf{q}}_{i,t}}(\cdot)$ is the dual quaternion-based transform. ${{\bf{R}}_{i,t}} \in SO(2)$ and ${{\bf{t}}_{i,t}} \in \mathbb{R}^2$ are the related rotation matrix and translational vector respectively, which are not explicitly used in our method.

Similarly, we denote the warp function on a pixel $p$ at time $t$ as $W_{p,t}=s_{p,t}{{\bf{q}}_{p,t}}$. In our system, we use the linear combination of the neighboring nodes to obtain $W_{p,t}$, that is

\begin{equation}
s_{p,t} = \sum_{i}^{N}\left ( w_i^p s_{i,t} \right ) / \sum_{i}^{N}w_i^p,
\label{eq_node2pixel1}
\end{equation}
and
\begin{equation}
{\bf{q}}_{p,t} = \sum_{i}^{N}\left ( w_i^p {\bf{q}}_{i,t} \right ) / \sum_{i}^{N}w_i^p,
\label{eq_node2pixel2}
\end{equation}

\noindent where $w_i^p$ is the weight between node $i$ and pixel $p$, which is determined by the distance between node $i$ and pixel $p$ at time 0, that is

\begin{equation}
w_i^p = \exp(-\alpha\left \| {\bf{x}}_{i,0} - {\bf{x}}_{p,0} \right \|^2)
\label{eq_ARAPweight}
\end{equation}

\noindent where $\alpha \in \mathbb{R}$ is a coefficient.

Hence, the deformation recovery problem is equivalent to the estimation of the warp functions $W_{i,t}$, $i=1,2,...,N$. We propose a novel 2D non-rigid SLAM framework based on EMDQ, as shown in Fig. \ref{fig_system}. This non-rigid SLAM system tracks the deformation from adjacent video frames using the EMDQ algorithm, which uses the SURF matching results as the input. However, this tracking strategy may result in accumulative errors and to solve this problem, a loop closing method is integrated into the system. The tracking and loop closing results are merged according to the uncertainties. This system also include methods to determine, propagate and merge the uncertainties. Finally, according to the deformation estimation results, we warp the coordinates of image pixels from time $t$ to $0$ according to Eq. \eqref{eq_node2pixel1} and \eqref{eq_node2pixel2}, and generate a larger mosaic in real-time by using CPU or GPU parallel computing technologies.

\subsection{Tracking}

For each video frame at time step $t$, we extract the SURF feature points and match them with the $t-1$ frame. The changes of $W_i$ for node $i=1,2,...N$ can be obtained from the EMDQ algorithm, which are denoted as $\Delta W_{i,t-1 \to t} = \Delta s_{i,t-1 \to t} \Delta {\bf{q}}_{i,t-1 \to t}$. Then, we update the warp functions of all node $i=1,2,...,N$ with

\begin{equation}
 W_{i,t, \text{track}} = f \left ( W_{i,t-1}, \Delta W_{i,t-1 \to t} \right ),
\label{eq_coordtrack}
\end{equation}

\noindent where $f(\cdot)$ is the updating function of the warp functions, which involves dual quaternion-based computation and the details are given in Appendix A. We use the subscript label "$\text{track}$" to denote the results of the tracking module.

\subsection{Loop Closing and Key Frames}

The results of the tracking module are accurate if the video sequence is short but may have accumulative errors for long video sequences. Hence, we integrate the loop closing module. As the example shown in Fig. \ref{fig_systemexample}, the basic idea of the loop closing method is to match the current video frame at time $t$ with the previous key frames, obtain the EMDQ deformation field with respect to the key frames, and then merge the tracking and loop closing results according to the uncertainties.

\noindent \textbf{Add new key frames:} The non-rigid SLAM system maintains a list of key frames in the computer memory for loop closing. We will add frame $t$ as a new key frame if the nodes have large displacements compared with those of the existing key frames, that is

\begin{equation}
\min_{k \in \Omega } \frac{1}{N}\sum_{i=1}^{N} \left \| {\bf{x}}_{i,t} - {\bf{x}}_{i,k} \right \| > H,
\label{eq_distanceFrame}
\end{equation}

\noindent where $k$ is the index of previous key frames and $\Omega$ is the set of key frame indexes, ${\bf{x}}_{i,t}$ and ${\bf{x}}_{i,k}$ are the coordinates of node $i$ at frame $t$ and key frame $k$ respectively, $H \in \mathbb{R}$ is a threshold. The information stored in the computer memory mainly include the SURF features and the warp functions of the deformation nodes.

\noindent \textbf{Select previous key frames for loop closing:} From the list of key frames, we select key frames that are close to the current frame $t$ for loop closing, and the distance metric is the same as in Eq. \eqref{eq_distanceFrame}. Using brute force, we compare all existing key frames with the tracking results ${\bf{x}}_{i,t,\text{track}}$, which are obtained by Eq. \eqref{eq_coordtrack} and \eqref{eq_warp}. This brute force search is effective because the number of key frames is often small for the following reasons: (1) the surgical scene is usually much smaller than that of rigid SLAM methods, hence it does not need a long video sequence for mosaicking, and (2) the method to add key frames, given by Eq. \eqref{eq_distanceFrame}, guarantees that the key frames have small overlap. For situations when long video sequences are required, such as fetoscopy or cystoscopy images, one may use the kd-tree structure for searching close frames. With a selected key frame $k$, we perform the SURF matching, the EMDQ computation and update the warp function \eqref{eq_coordtrack} to obtain the estimation results of the warp functions at time $t$, that is

\begin{equation}
W_{i,t,\text{loop}} = f \left ( W_{i,k}, \Delta W_{i,k \to t} \right ),
\label{eq_coordloop}
\end{equation}

\noindent where $\Delta W_{i,k \to t} = \Delta s_{i,k \to t}\Delta {\bf{q}}_{i,k \to t}$ is the change in warp function of node $i$ from key frame $k$ to frame $t$ obtained by the EMDQ algorithm.

\noindent \textbf{Tracking Failure:} Too fast laparoscope motion and/or large glossy components of the tissue surface may cause a failure in the feature tracking process. In traditional rigid SLAM systems, this problem is usually solved in the loop closing module by matching the previous key frames. For short video sequences, the number of previous key frames is limited and it is feasible use brute force search. For long video sequences, one may refer to the bag-of-words (BoW) method that is widely used in large-scale rigid SLAM systems for obtaining candidate key frames to match the current frame \cite{mur2015orb}. In addition to the technology solution, it is also possible to include the surgeon in the image mosaicking process by warning too fast laparoscope motion to retrieve the tracking, since our method works in real-time.

\subsection{Uncertainty}

The estimated warp functions from the tracking and loop closing modules are $W_{i,t,\text{track}}$ and $W_{i,t,\text{loop}}$ respectively, which may be different and we merge them according to the uncertainties. The basic idea is borrowed from the extended Kalman filter (EKF) that uses Gaussian distributions to assign soft weights for merging.

\noindent \textbf{Uncertainty of the EMDQ results:} According to Eq. \eqref{eq_coordtrack} and Eq. \eqref{eq_coordloop}, the warp functions are updated according to results of EMDQ. Hence, the uncertainty of the EMDQ results is essential for estimating the uncertainty of the tracking and loop closing results. Because the EMDQ algorithm generates the deformation field by interpolating among the inliers of the feature matches, it is intuitive that if node $i$ is distant from the inliers of feature matches, the uncertainty of $\Delta W_i = \Delta s_i\Delta {\bf{q}}_i$ is high. Under this analysis, the uncertainty of $\Delta W_i$ is

\begin{equation}
\Delta \sigma_{i}^2 = \min_j (\exp(\beta d_{i,j}^2)),
\label{eq_sigmaEMDQ}
\end{equation}

\noindent where $\beta$ is a coefficient, $d_{i,j}$ is the distance between node $i$ and the feature inlier $j$. For each node $i=1,2,..N$, we search all feature inliers and use the ones that provide the minimum uncertainty. We use the exponential function to make the uncertainty small at areas that are close to the features, and increase significantly at distant areas. This design is consistent with the observation that the estimation of areas that are close to the control points are much more certain and accurate.

\noindent \textbf{Uncertainty propagation:} According to Eq. \eqref{eq_coordtrack}, the uncertainty of node $i$ at time $t$, $\sigma_{i,t}^2$, should be updated from $\sigma_{i,t-1}^2$ and $\Delta \sigma_{i,t-1 \to t}^2$. Because $W_{i,t-1}$ is independent with $\Delta W_{i,t-1 \to t}$, the uncertainty of node $i$ is propagated by

\begin{equation}
\sigma_{i,t,{\text{track}}}^{2} = \Delta s_{i,t-1 \to t}^2 \sigma_{i,t-1}^{2} + \Delta \sigma_{i,t-1 \to t}^2.
\label{eq_sigmatrack}
\end{equation}

The uncertainties of the loop closing results, $\sigma_{i,t,{\text{loop}}}^{2}$, are obtained similarly.

However, because the tracking module updates the warp functions from each two adjacent frames, that is $t \to t+1 \to t+2 \to...$, the uncertainties of nodes may increase too fast according to the prorogation method \eqref{eq_sigmatrack}, since $\Delta \sigma_{i,t-1 \to t}^2$ is determined by the distance between node $i$ and the image features (see \eqref{eq_sigmaEMDQ}). To solve this problem, we propose to take into account the feature matching relationships among multiple frames. For example, for a feature point that can be tracked continuously at multiple frames, the uncertainties of its neighboring nodes should not increase too fast. Hence, we introduce the concepts of feature uncertainty. For a feature $j$ at time $t$, if it is an inlier (can find the correct correspondence at $t-1$ by EMDQ), then its uncertainty is propagated by

\begin{equation}
\sigma_{j,t,{\text{feature}}}^{2} = \sigma_{j,t-1,{\text{feature}}}^{2} + \sigma_{j,\text{EMDQ}}^2,
\label{eq_sigmafeaturepropagate}
\end{equation}

\noindent where $\sigma_{j,\text{EMDQ}}^2$ is the squared error of feature $j$ when performing the EMDQ algorithm between $t-1$ and $t$. Because only matches with small errors are considered as inliers by the EMDQ algorithm, $\sigma_{j,\text{EMDQ}}^2$ is small and the increasing rate of the uncertainties of feature inliers is small. If feature $j$ is an outlier at time $t$, then $\sigma_{j,t,{\text{feature}}}^{2}$ is determined in the same way as in Eq. \eqref{eq_sigmatrack}.

Then, we introduce a spatial restriction between the uncertainties of nodes and features in the tracking module. For a node $i$, its uncertainty should satisfy

\begin{equation}
\sigma_{i,t,\text{track}}^2 \le \sigma_{j,t,{\text{feature}}}^{2} + \exp(\beta d_{i,j}^2),
\end{equation}

\noindent for any image feature $j$. Because the increase in the feature uncertainties are limited by \eqref{eq_sigmafeaturepropagate}, the increase of node uncertainties can also be limited.

\noindent \textbf{Uncertainty-based Merging:} For each node $i=1,2,...N$, we consider the tracking and loop closing modules as two sensors, and merge their results $W_{i,t,\text{track}}$ and $W_{i,t,\text{loop}}$ by using the extended Kalman filter (EKF) according to the uncertainties $\sigma_{i,t,{\text{track}}}^{2}$ and $\sigma_{i,t,{\text{loop}}}^{2}$. The following merging algorithm is adapted from Ref.\cite{sun2004multi}. For node $i$, the covariance matrix is

\begin{equation}
{\bf{A}}_i = \begin{bmatrix}
\sigma_{i,t,{\text{track}}}^{2} & \eta_i \sigma_{i,t,{\text{track}}}\sigma_{i,t,{\text{loop}}}\\
\eta_i \sigma_{i,t,{\text{track}}}\sigma_{i,t,{\text{loop}}} & \sigma_{i,t,{\text{loop}}}^{2}
\end{bmatrix},
\label{eq_A}
\end{equation}

\noindent where $\eta_i \in [0.0, 1.0]$ is the correlation coefficient, which is used in EKF to suggest the correlation relationship between sensors. In this system we determine $\eta$ by

\begin{equation}
\eta_i = \exp(-\gamma \left \| {\bf{x}}_{i,t} - {\bf{x}}_{i,k} \right \|^2),
\label{eq_eta}
\end{equation}

\noindent where ${\bf{x}}_{i,t}$ and ${\bf{x}}_{i,k}$ are the coordinates of node $i$ at time $t$ and key frame $k$ respectively, $\gamma$ is a coefficient. Then, the merged uncertainty is

\begin{equation}
\sigma_{i,t,\text{merge}}^2=1/\sum \left ( {\bf{A}}_i^{-1} \right ),
\label{eq_sigmamerge}
\end{equation}

\noindent and the weights of the two sensors are

\begin{equation}
{\bf{w}}_i = \sigma_{i,t,\text{merge}}^2{\bf{A}}_i^{-1} \begin{bmatrix}
1\\
1
\end{bmatrix}.
\label{eq_mergeweight}
\end{equation}

According to the weights \eqref{eq_mergeweight}, we take the weighted average of $W_{i,t,\text{track}}$ and $W_{i,t,\text{loop}}$ and obtain a new warp function $W_{i,t,\text{merge}}$ for node $i$ at time $t$. However in practice we found that the equations \eqref{eq_A} to \eqref{eq_mergeweight} may result in negative weights ${\bf{w}}_i$ if $\eta_i$ is too large, in that case we will simply take the values related to the smaller uncertainty as the merged values.

The uncertainties increase as in the tracking module (Eq. \eqref{eq_sigmatrack}), and decrease after merging with the loop closing results (Eq. \eqref{eq_sigmamerge}). In this way our system maintains the uncertainties of nodes at a low level.

The above EKF-based merging is equivalent to linear merging if $\eta_i = 0$. EKF-based merging is more appropriate to handle slow motion when the results of tracking and loop closing modules are obtained from very close frames, hence one of them should be omitted without decreasing the uncertainty by Eq. \eqref{eq_sigmamerge}. When the motion is slow, $\eta_i \approx 1$ according to \eqref{eq_eta}, which will result in the omission.

\subsection{ARAP-based Smoothing}

After merging the results of tracking and loop closing modules, we add an as-rigid-as-possible (ARAP) smoothing \cite{sorkine2007rigid} step to obtain the final estimation results of the warp functions at time $t$. ARAP smoothing is widely used in the non-rigid SLAM systems \cite{newcombe2015dynamicfusion}, which is usually integrated into the cost function and minimized by a Gauss-Newton-like optimization method. Because these real-time optimization methods often run on a powerful GPU, we propose a novel iterative method that uses closed form ARAP results to update the warp functions, which is also effective on the CPU. Specifically, the ARAP warp function of node $i$ at time $t$, $W_{\text{ARAP}}$, is computed from the change of the coordinates of its neighboring nodes between time 0 and $t$. Specifically,

\begin{equation}
{\bf{C}}_0 = \begin{bmatrix}
{w_1^i(\bf{x}}_{1,0} - {\bf{x}}_{i,0}) & ... & w_N^i({\bf{x}}_{N,0} - {\bf{x}}_{i,0})
\end{bmatrix}_{2 \times N},
\end{equation}

\begin{equation}
{\bf{C}}_t = \begin{bmatrix}
{w_1^i(\bf{x}}_{1,t} - {\bf{x}}_{i,t}) & ... & w_N^i({\bf{x}}_{N,t} - {\bf{x}}_{i,t})
\end{bmatrix}_{2 \times N},
\end{equation}

\noindent where $w_j^i = \exp(-\alpha \left \| {\bf{x}}_{i,0} - {\bf{x}}_{j,0} \right \|^2)$ is the weight between node $i$ and $j$, which is similar to Eq. \eqref{eq_ARAPweight}. In practice we will remove the related columns of ${\bf{C}}_0$ and ${\bf{C}}_t$ if $w_j^i$ is too small for faster computation. It is worth noting that there exists a trade-off between speed and robustness, because a large $\alpha$ will reduce the number of neighboring nodes for ARAP smoothing, which will result in faster speed but lower robustness.

Then, following Ref.\cite{arun1987least}, the rotation matrix ${\bf{R}}_{\text{ARAP}}$, translation vector ${\bf{t}}_{\text{ARAP}}$ and scale ${s}_{\text{ARAP}}$ can be obtained by

\begin{equation}
[{\bf{U}}, \Sigma, {\bf{V}}^T] = \text{svd}({\bf{C}}_t{\bf{C}}_0^T),
{\bf{R}}_{\text{ARAP}}= {\bf{U}}{\bf{V}}^T
\label{eq_ARAP1}
\end{equation}

\begin{equation}
s_{\text{ARAP}} = \left \| \text{vector}({\bf{C}}_t) \right \| / \left \| \text{vector}({\bf{C}}_0) \right \|
\label{eq_ARAP2}
\end{equation}

\begin{equation}
t_{\text{ARAP}} = \text{weighted average}({\bf{X}}_{t} / s_{\text{ARAP}} - {\bf{R}}_{\text{ARAP}}{\bf{X}}_{0})
\label{eq_ARAP3}
\end{equation}

Then we generate the quaternion from ${\bf{R}}_{\text{ARAP}}$ and $t_{\text{ARAP}}$.

Denoting the ARAP warp functions of node $i$ at time $t$ as $W_{i,t,\text{ARAP}} = \left \{ s_i^t, {{\bf{q}}_i^t} \right \}_{\text{ARAP}}$, our goal is obtain the warp function $W_{i,t} = s_{i,t}{\bf{q}}_{i,t}$ that minimizes

\begin{equation}
\left \| W_{i,t,\text{merge}}({\bf{x}}_{i,0})- W_{i,t}({\bf{x}}_{i,0}) \right \|^2 + \lambda\left \| W_{i,t,\text{ARAP}}({\bf{x}}_{i,0})- W_{i,t}({\bf{x}}_{i,0}) \right \|^2.
\label{eq_ARAPcost}
\end{equation}

\noindent where $\lambda = (1 + \sigma_{i,t}^2)/(1 + \sigma_{\text{ARAP}}^2)$ is a coefficient suggesting the weight of the ARAP term, $\sigma_{\text{ARAP}}^2 = 100$ is a fixed value. In this cost function, $W_{i,t,\text{merge}}$ is the data term, which is determined by the merging method and is fixed in this step. We update $W_{i,t}$ by

\begin{equation}
W_{i,t} = (W_{i,t,\text{merge}} + \lambda W_{i,t,\text{ARAP}}) / (1 + \lambda).
\label{eq_ARAPupdate}
\end{equation}

Our method to minimize cost \eqref{eq_ARAPcost} is to iteratively estimate $W_{i,t,\text{ARAP}}$ by using Eq. \eqref{eq_ARAP1}-\eqref{eq_ARAP3}, and then estimate the new $W_{i,t}$ by using Eq.\eqref{eq_ARAPupdate} to update ${\bf{x}}_{i,t}$ according to Eq. \eqref{eq_warp}. We check the cost \eqref{eq_ARAPcost} after each iteration, and will terminate the process if the cost increases. In practice we found that a few iterations can obtain good results hence we set the maximum number of iterations to 5.

\begin{figure} [htp]
\vspace{0.0cm}
\centering
  \includegraphics[width=0.45\textwidth]{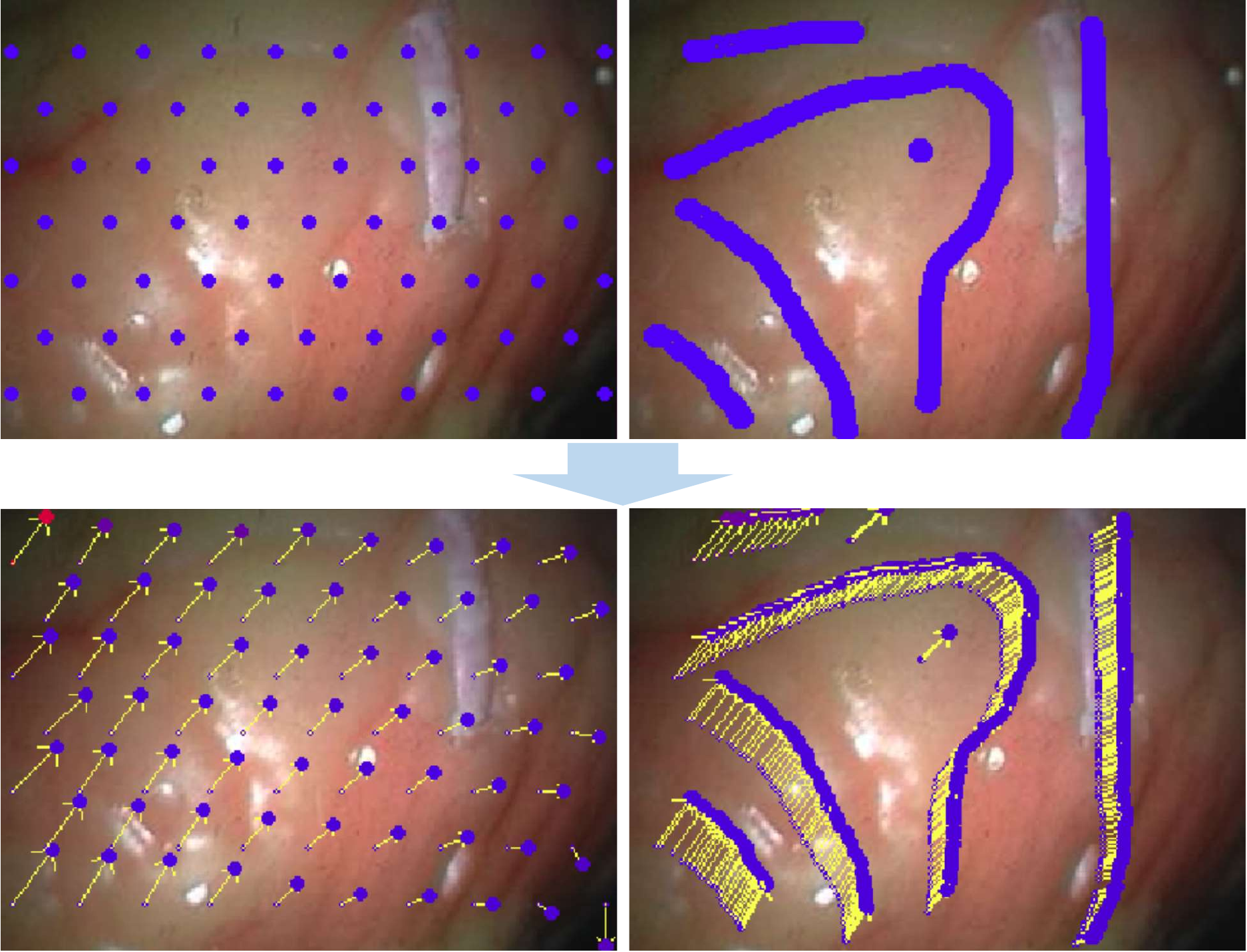}
  \caption{An example to demonstrate the non-rigid pixel tracking ability with a heart phantom that simulated heartbeat. Left images: the tracking results of nodes. Right images: tracking results of other pixels, which are manually drawn on the upper-right image and tracked to other frames by interpolating among the nodes. The arrows are the displacements of nodes or pixels.}
\label{fig_pixeltracking3}
\end{figure}

As the example shown in Fig. \ref{fig_pixeltracking3}, the initial deformation nodes can be tracked robustly by using the methods introduced in sections III.A-III.D. In addition, all pixels can also be tracked according to the interpolation among the nodes, which is a prerequisite for dense and non-rigid image mosaicking.

\subsection{Adding New Nodes}

At time 0, we insert the first node at the center of the image, and then insert new nodes by checking the 6-neighboring locations of existing nodes iteratively until all image areas are covered. The same node inserting strategy is also performed when new observed areas are distant from the existing nodes. The warp functions of new nodes are equal to the weighted average value of the neighboring existing nodes, where the weights are the same as the ARAP weights \eqref{eq_ARAPweight}.

\subsection{Dense and Non-rigid Image Mosaicking in Real-time}

As the reference frame, frame $0$ is inserted directly to the mosaicking image. At time $t$, we compute the warp functions of nodes $W_t = \left \{ {W_{1,t},W_{2,t},...,W_{N,t}} \right \}$, and then compute the warp effects on each pixel by interpolating among the neighboring nodes following Eqs. \eqref{eq_node2pixel1} and \eqref{eq_node2pixel2}. The coordinate of pixel $p$ at time $0$ can be obtained by ${\bf{x}}_{p,0} = W_p^{-1}({\bf{x}}_{p,t})$, and the related RGB value, $\text{rgb}_t$, will be merged to the mosaicking image at ${\bf{x}}_{p,0}$. To make the mosaicking image more smooth, we have developed the truncated signed distance function (TSDF)-like \cite{curless1996volumetric} method to merge frame $t$ with the large mosaicking image, that is

\begin{equation}
{\text{rgb}}_{\text{merge}} = w_{\text{merge}}{\text{rgb}}_{\text{merge}} + {\text{rgb}}_{t},
\label{eq_TSDF1}
\end{equation}

\begin{equation}
w_{\text{merge}} = \max   \left \{ w_{\text{merge}} + 1, 30 \right \},
\label{eq_TSDF2}
\end{equation}

The above computations, including both ${\bf{x}}_{p,0}$ estimation and RGB values merging, need to be performed for all pixels, which is computationally expensive because the laparoscopic images may have millions of pixels. Note that the computations are independent for each pixel, hence parallel computational technologies can be used for acceleration. We have developed both CPU and GPU-based parallel computation methods. The CPU parallel computation is based on the OpenMP library to make full use of all CPU cores, and the GPU parallel computation is developed using CUDA C++, which launches GPU threads for each pixel in parallel. In practice we found that a powerful CPU can achieve the real-time requirement, while the GPU-based computation can make the system performance 2x faster than the CPU version.

\subsection{Parameters Setting}

Key parameters used in our method are as follows: $\alpha=$ 2e-4 is used to compute the weight between nodes and pixels (see Eq. \eqref{eq_ARAPweight}), which is also used for computing the ARAP weights between nodes. $\beta=$ 3e-3 is used to compute the uncertainty of nodes according to the distance to feature inliers (see Eq. \eqref{eq_sigmaEMDQ}). $\gamma=$ 5e-3 is used to correlation coefficient for uncertainty-based merging (see Eq. \eqref{eq_eta}). Every $K=5$ frames, we perform the loop closing step and every 2 frames are used for mosaicking. Because $\alpha$, $\beta$ and $\gamma$ are related to squared distances in pixels, to make the algorithms self-adaptive to different image resolutions, we propose to adjust $\alpha$, $\beta$ and $\gamma$ by multiplying $1/s^2$. By considering $480\times 270$ as the reference resolution, $s = (w / 480 + h / 270)/2$ suggests the scale change of the input video frames, where $w$ and $h$ are the width and height of the video frames.

\section{Experiments}

\begin{figure*} [htp]
\vspace{0.0cm}
\centering
  \includegraphics[width=0.95\textwidth]{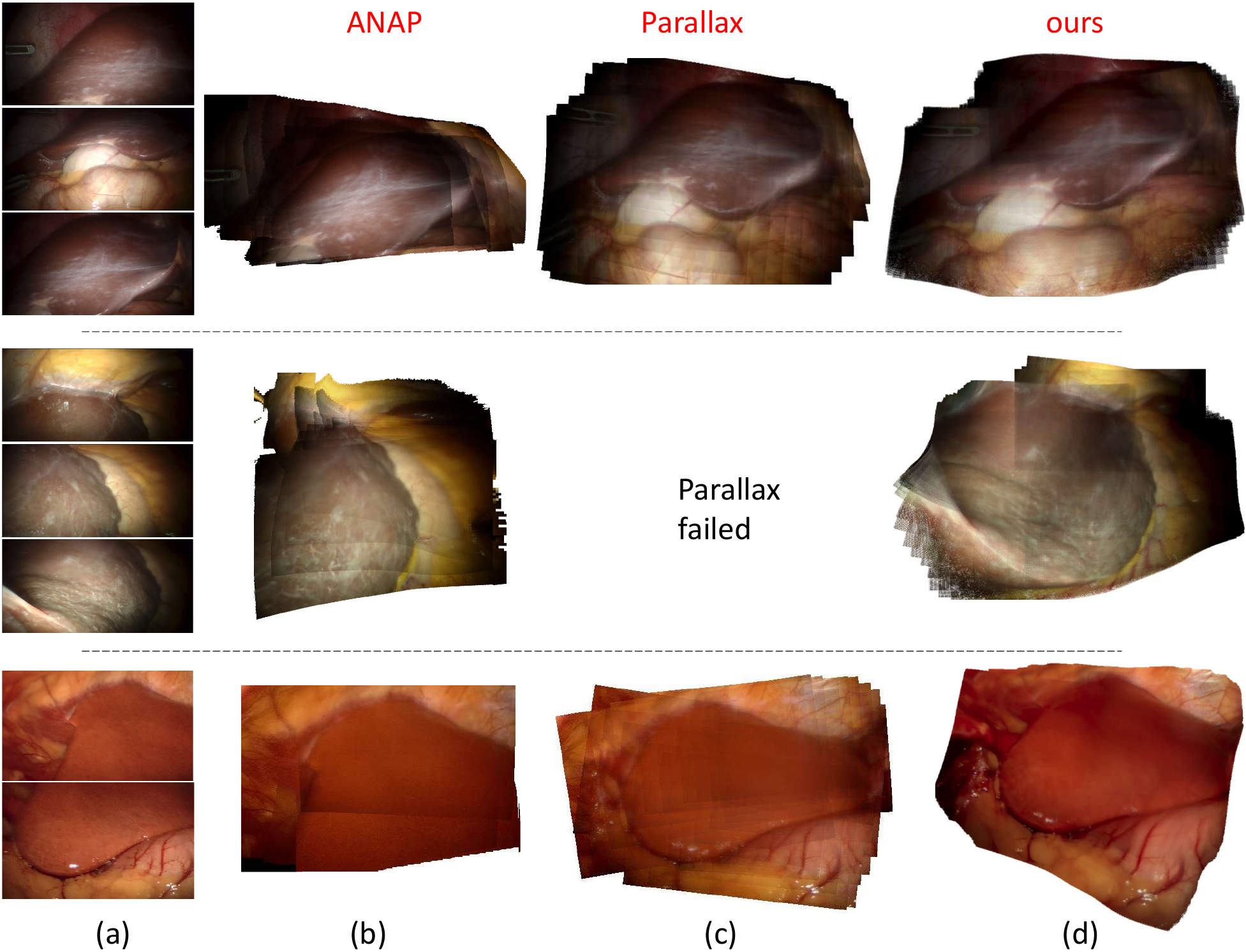}
  \caption{Experiments on laparoscopy videos captured during a robot-assisted liver surgeries at our hospital. The liver had small deformation due to respiration. (a) Sample frames. (b) ANAP. (c) Parallax. (d) Ours. Number of frames: 161, 132 and 844 respectively. Resolution:  $728\times 392$, $728\times 392$ and $440\times 280$ respectively. Average GPU/CPU computational time per frame: 93.0/158.4, 98.0/189.1 and and 58.1/79.0 ms respectively. The number of down sampled images for ANAP and Parallax were 30, 35 and 20 respectively, and the total ANAP/Parallax runtime for the three cases were 5/23, 6/35 and 3/20 minutes respectively.}
\label{fig_exp345}
\end{figure*}

\begin{figure*} [htp]
\vspace{0.0cm}
\centering
  \includegraphics[width=0.95\textwidth]{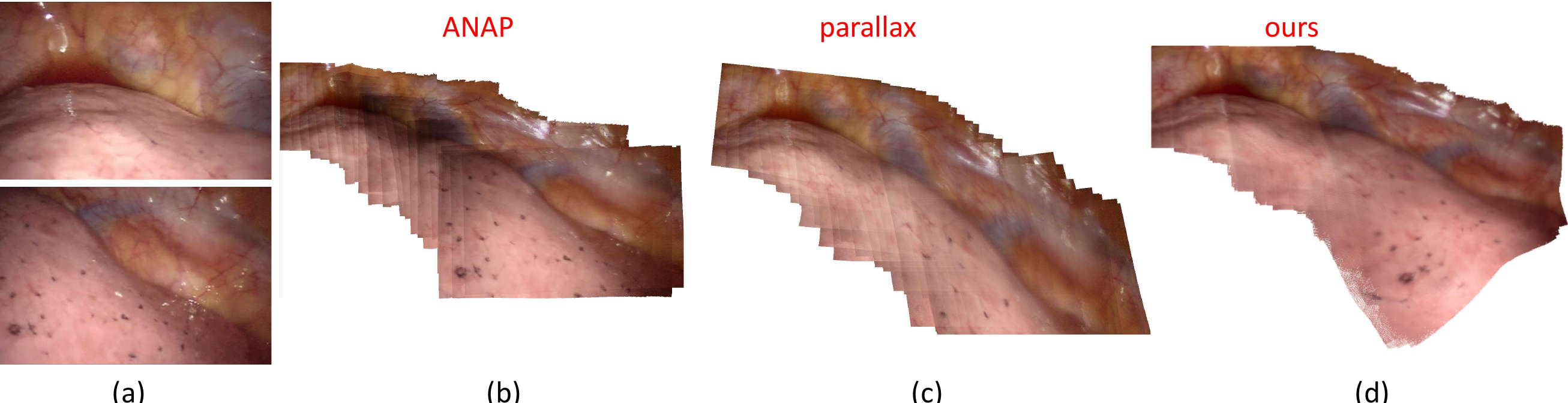}
  \caption{Experiments on laparoscopy videos captured during minimally invasive lung surgery at our hospital. The lung has significant deformation due to heartbeat. (a) Sample frames. (b) ANAP. (c) Parallax. (d) Ours. Number of frames: 227. Resolution: $484\times 312$. Average GPU/CPU computational time per frame: 59.2/106.2 ms. The number of down sampled images for ANAP and Parallax was 28, and the total ANAP/Parallax runtime were 4/16 minutes.}
\label{fig_exp5}
\end{figure*}

\begin{figure*} [htp]
\vspace{0.0cm}
  \includegraphics[width=0.95\textwidth]{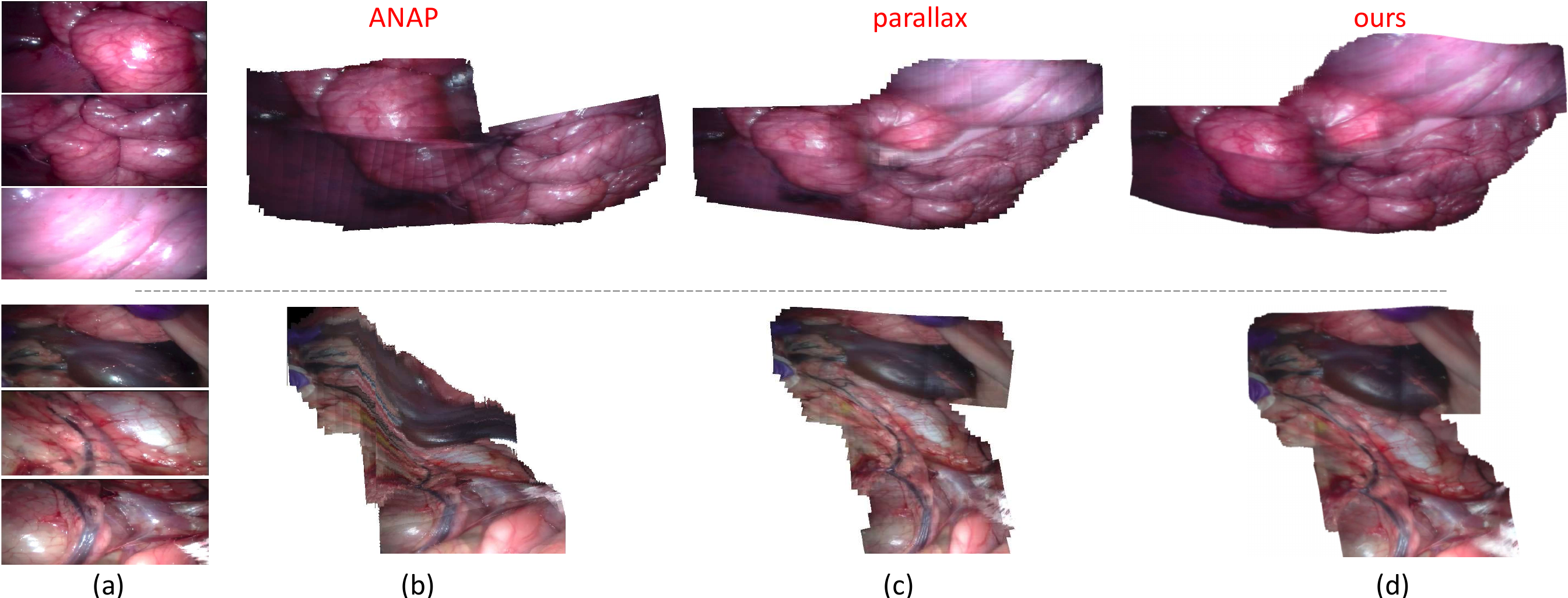}
  \caption{Experiments on the Hamlyn dataset. (a) Sample frames. (b) ANAP. (c) Parallax. (d) Ours. Number of frames: 387 and 362 respectively. Resolution:  $680\times 248$ for both. Average GPU/CPU computational time per frame: 61.3/70.2 ms and 71.6/120.4 ms respectively. The number of down sampled images for ANAP and Parallax was 67 and 43 respectively, and the total ANAP/Parallax runtime were 9/46 and 7/33 minutes respectively.}
\label{fig_exp78}
\end{figure*}

\begin{figure*} [htp]
\vspace{0.0cm}
  \includegraphics[width=0.95\textwidth]{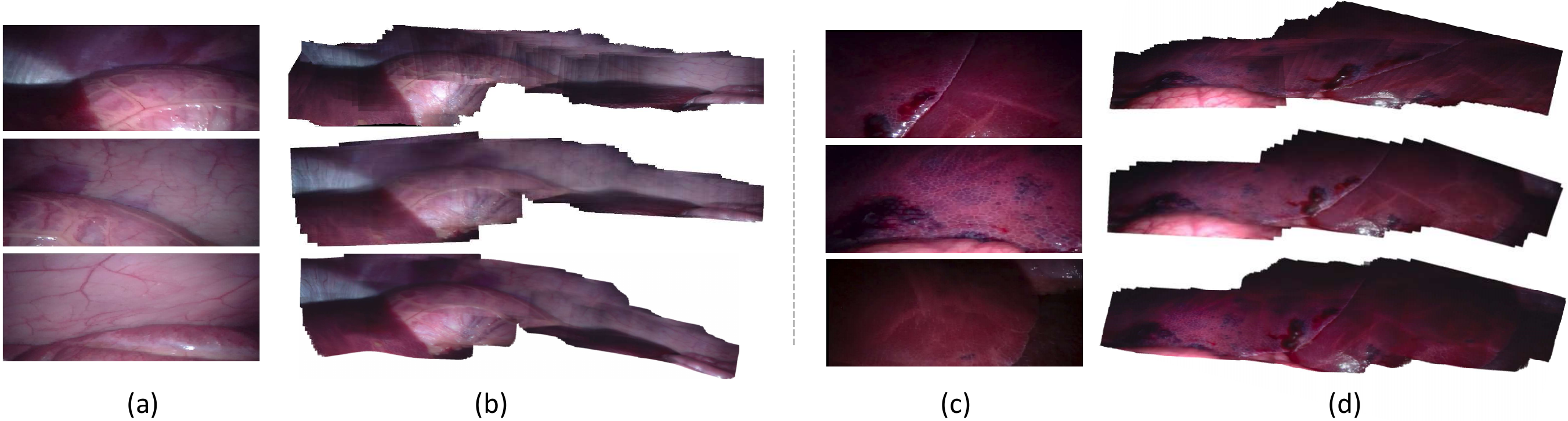}
  \caption{Experiments on the Hamlyn dataset, which include long image sequences. (a) and (c) are the sample images of the two cases respectively. From top to bottom in (b)-(d) are the results of ANAP, Parallax and ours. Number of frames: 741 and 772 respectively. Resolution:  $680\times 248$ for both. Average GPU/CPU computational time per frame: 80.0/142.9 ms and 69.3/119.0 ms respectively. The number of down sampled images for ANAP and Parallax was 80 and 63 respectively, and the total ANAP/Parallax runtime were 11/52 and 8/40 minutes respectively.}
\label{fig_exp9}
\end{figure*}

The source code was implemented in C++ and ran on a desktop with an Intel Core i9 3.0 GHz CPU (16 cores) and NIVIDA Titan RTX GPU.

To evaluate the performance of the proposed mosaicking method on laparoscopic images in real-world surgical scenarios, we obtained intraoperative videos during surgeries performed in our hospital and online videos \footnote{http://hamlyn.doc.ic.ac.uk/vision/}. The videos were recorded under an Institution Review Board approved protocol. We compared the results of our algorithm to the the as-natural-as-possible (ANAP) \cite{lin2015adaptive} and Parallax \cite{zhang2014parallax} image mosaicking methods for comparison. Both ANAP and Parallax include elastic warping mechanisms to handle image deformation. Due to the heavy computational burden, both ANAP and Parallax are off-line methods and their codes were available as Matlab scripts. In our experiments, we down-sampled the video frames to reduce the computational time of ANAP and Parallax since they are very slow for long image sequences. For the runtime, we report the average per frame computational time for our method since it is an online real-time method, and report the total computational time for ANAP and Parallax since they are off-line methods.

As shown in Fig. \ref{fig_exp345}, the first experiments were conducted on laparoscopy videos captured during robot-assisted liver surgery. We asked the surgeons to move the laparoscope within the patients' abdomens. The livers had small deformation caused by respiration. The translational motion of the laparoscope is large and the tissue surfaces had significant 3D shapes, which caused large pixel deformation due to parallax. For these cases, ANAP and Parallax are not as robust as our method.

For the experiment shown in Fig. \ref{fig_exp5}, the videos were captured during a minimally invasive sublobar lung surgery. Due to heartbeat and respiratory motion caused by the adjacent lung, the deflated lung had significant and fast deformation. For this data, the camera motion was mainly along the tissue surface without significant changes in magnification, which made it easier to match adjacent video frames. Hence, all three methods were able to obtain satisfying results.

In the experiments shown in Fig. \ref{fig_exp78} and \ref{fig_exp9}, the laparoscopy videos were obtained from the Hamlyn public dataset, which includes videos showing the porcine abdomen deformed due to respiration. Our method and the Parallax method can obtain good mosaicking results.

\begin{figure*} [t]
\vspace{0.0cm}
\centering
  \includegraphics[width=1.0\textwidth]{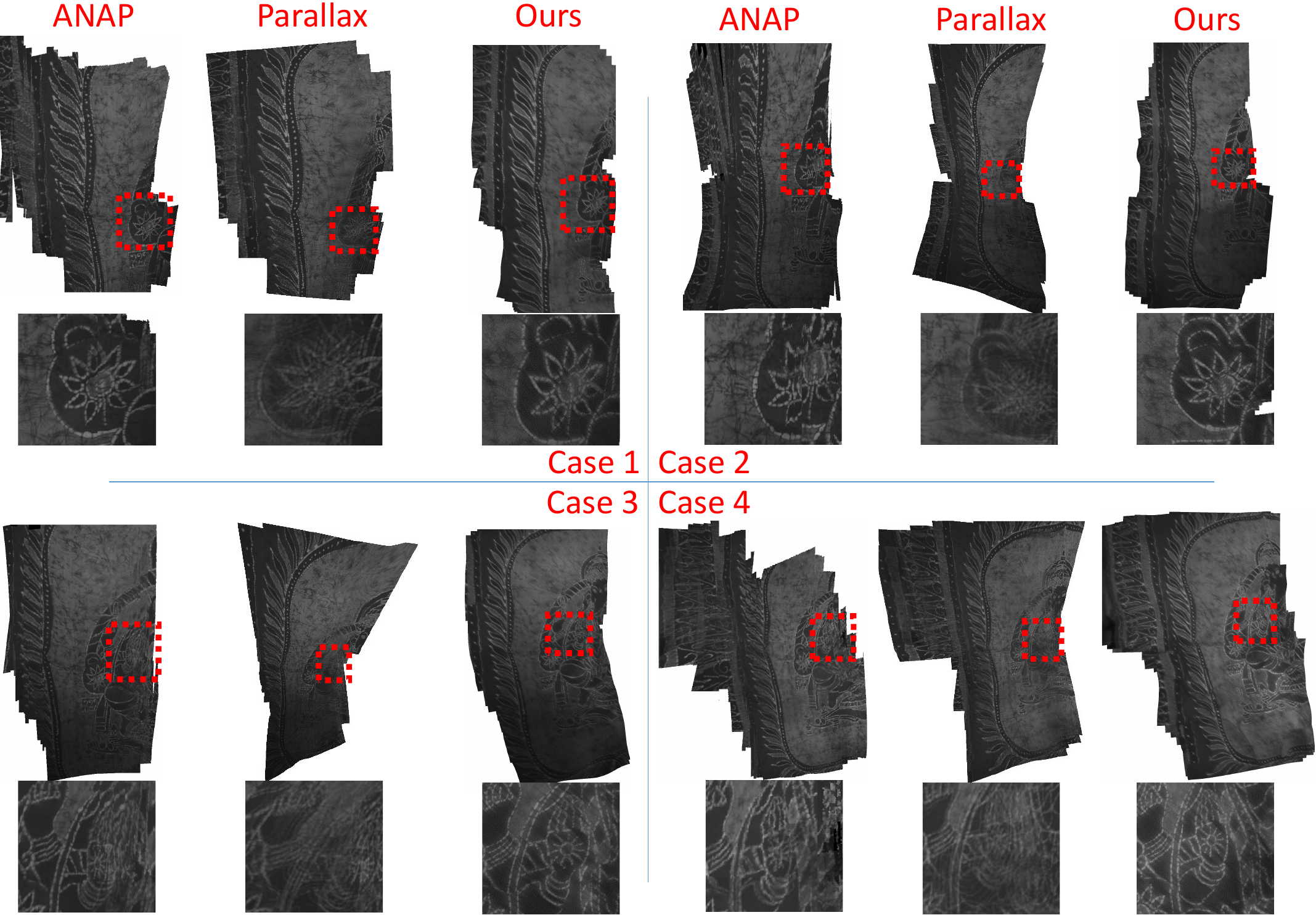}
  \caption{Experiments on the Mandala data. Deformation increased from case 1 to case 4. For each case, up: the mosaicking results, bottom: parts within the red dash lines are enlarged for better visualization of the texture details. Our method demonstrated much better accuracy.}
\label{fig_expmandala}
\end{figure*}

\begin{figure} [t]
\vspace{0.0cm}
\centering
  \includegraphics[width=0.45\textwidth]{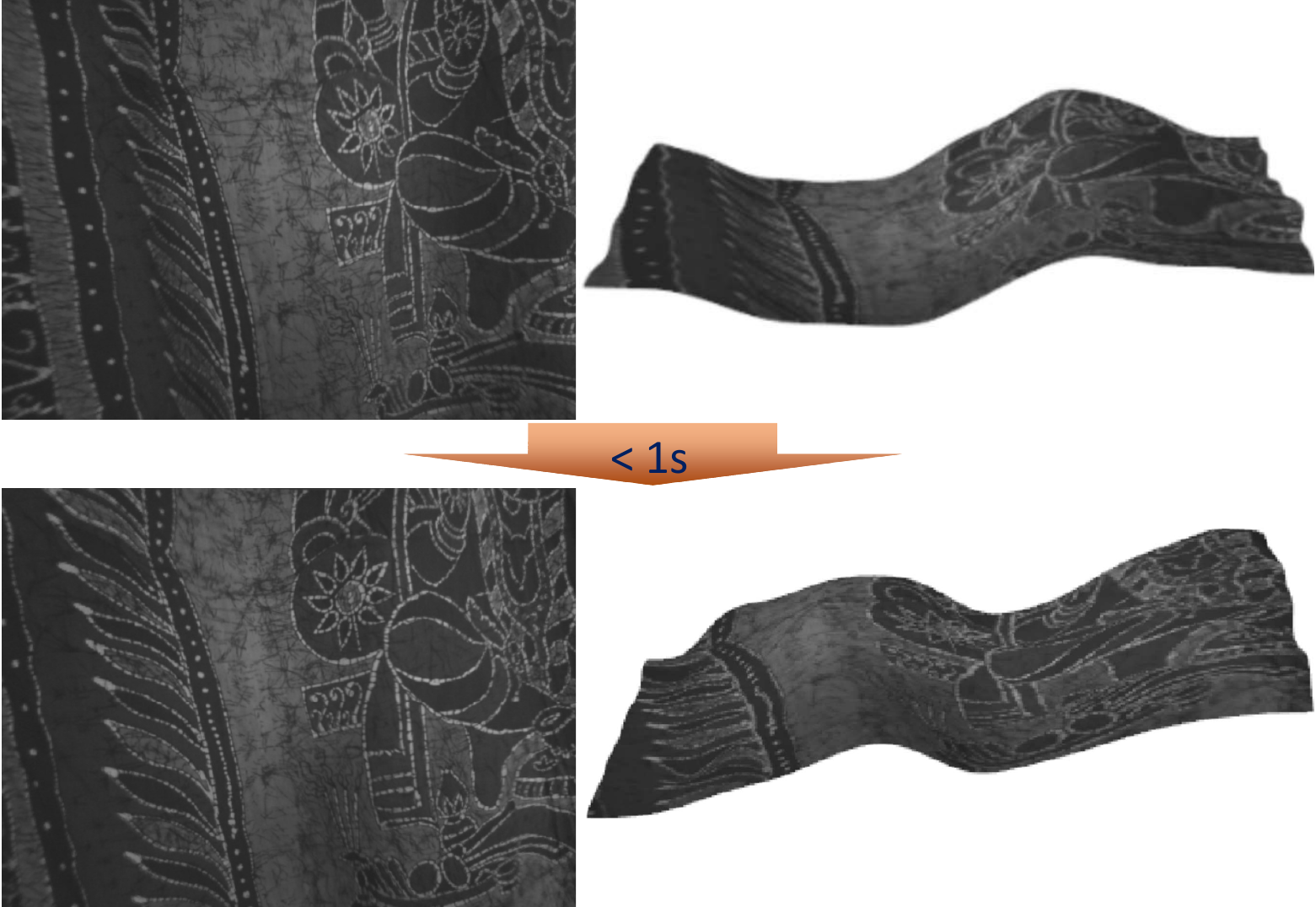}
  \caption{Left: sample images of case 4 in the Mandala dataset. Right: side view of the related 3D template model generated by a stereo matching method \cite{zhou2019real}, which were rendered by VTK. Note that the 3D model is not needed for 2D mosaicking. The elapsed time between the two sample images was less than one second, which suggests the deformation is large and fast.}
\label{fig_expmandalaSample}
\end{figure}

In these above experiments on \textit{in vivo} laparoscopy videos, our online real-time mosaicking method obtained excellent results, which were comparable with the results of the off-line methods. Since the tissue deformation in most of the collected \textit{in vivo} videos were relatively small, the ability to handle large deformation was not fully evaluated. Hence, we introduce the Mandala dataset from Ref. \cite{lamarca2020defslam}, which includes four gray-level image sequences of a soft blanket. As shown in Fig. \ref{fig_expmandalaSample}, the deformation of the blanket increased from case 1 to case 4, and the camera scanned the deforming blanket. The deformation in the Mandala data was significant and fast. Our method demonstrated better ability to handle large deformation to reserve more texture details, as illustrated in Fig. \ref{fig_expmandala}. In this experiments, the images blending method was the multi-band blending (MBB) method \cite{brown2007automatic}, which can handle larger registration errors caused by the large deformation. However, MBB may not be appropriate for blending laparoscopy images due to the complex illumination condition. Hence, for laparoscopy images mosaicking, we found the TSDF-based methods (Eq. \eqref{eq_TSDF1} and \eqref{eq_TSDF2}) can obtain better results.

\section{Conclusion}

We have developed a novel 2D non-rigid SLAM system for laparoscopy images mosaicking. To the best of our knowledge, this is the first image mosaicking method that can handle large translational camera motion, complex anatomical surfaces and tissue deformation in real-time, which makes it possible to mosaic \textit{in vivo} laparoscopic images. Experiments with real-world \textit{in vivo} data on different types of organs have shown the feasibility of our method. This would be particularly useful for minimally invasive surgeries wherein a larger field of view can provide the surgeon with a precise anatomical map to plan the surgery (for example segmentectomy) or placement of a device (meshes for hernias).

\noindent \textit{Limitations and future works}: (1) 2D image mosaicking has an intrinsic requirement that different images of 3D objects can be reprojected to the same mosaic, hence it may not be able to mosaic images obtained at arbitrary positions. Hence, our method is mainly designed for situations when the laparoscope moves along the tissue surface, and cannot handle some types of camera motions, such as going through the tubular airways/vessels. (2) Our method works under a reasonable assumption that during the scan, the instruments are removed from the field of view. Occlusion by surgical instruments may interfere with the SURF feature matching results and further affect the deformation field generated by EMDQ. Future works will include the integration of instrument segmentation to improve the robustness. (3) Because EMDQ assumes the deformation is smooth, our method cannot handle sharp non-rigid motions of the tissue, such as cutting of tissue surfaces.


%

\appendices
\section{Dual Quaternion-based Warp Function Updating}

The updating function of the warp function used in \eqref{eq_coordtrack} is
\begin{equation}
\left \{ s_{\text{new}}, {{\bf{q}}_{\text{new}}} \right \} = f \left ( s_{\text{old}}, {{\bf{q}}_{\text{old}}}, \Delta s, \Delta {\bf{q}} \right ),
\end{equation}

\noindent which suggest that if
\begin{equation}
x_1 =  s_{\text{old}} {{\bf{q}}_{\text{old}}}(x_0),
\end{equation}

\noindent then $s_{\text{new}}$ and ${{\bf{q}}_{\text{new}}}$ should have

\begin{equation}
s_{\text{new}} {{\bf{q}}_{\text{new}}}(x_0) = \Delta s \Delta {\bf{q}}(x_1).
\end{equation}

Under this requirements, the details of the computation are as follows:

\begin{equation}
{\bf{q}}_{1} = \text{trans2dq}\left ( (1-s_{\text{old}}) / s_{\text{old}} \Delta{\bf{t}} \right ),
\end{equation}

\noindent where $\text{trans2dq}(\cdot)$ is the function that convert a translational vector to the related dual quaternion. Then,

\begin{equation}
{\bf{q}}_{2} = \Delta {\bf{q}} * {\bf{q}}_{1},\\
\end{equation}

\begin{equation}
{\bf{q}}_{\text{new}} = {{\bf{q}}_{\text{old}}} * {\bf{q}}_{2} \text{ and } s_{\text{new}} = \Delta s s_{\text{old}},
\end{equation}

\noindent where $*$ is the multiply function between two dual quaternions.


%


\ifCLASSOPTIONcaptionsoff
  \newpage
\fi

\bibliographystyle{IEEEtran}
\bibliography{IEEEabrv,bare_jrnl}

\begin{thebibliography}{10}
\providecommand{\url}[1]{#1}
\csname url@samestyle\endcsname
\providecommand{\newblock}{\relax}
\providecommand{\bibinfo}[2]{#2}
\providecommand{\BIBentrySTDinterwordspacing}{\spaceskip=0pt\relax}
\providecommand{\BIBentryALTinterwordstretchfactor}{4}
\providecommand{\BIBentryALTinterwordspacing}{\spaceskip=\fontdimen2\font plus
\BIBentryALTinterwordstretchfactor\fontdimen3\font minus
  \fontdimen4\font\relax}
\providecommand{\BIBforeignlanguage}[2]{{%
\expandafter\ifx\csname l@#1\endcsname\relax
\typeout{** WARNING: IEEEtran.bst: No hyphenation pattern has been}%
\typeout{** loaded for the language `#1'. Using the pattern for}%
\typeout{** the default language instead.}%
\else
\language=\csname l@#1\endcsname
\fi
#2}}
\providecommand{\BIBdecl}{\relax}
\BIBdecl

\bibitem{bergen2014stitching}
T.~Bergen and T.~Wittenberg, ``Stitching and surface reconstruction from
  endoscopic image sequences a review of applications and methods,'' \emph{IEEE
  Journal of Biomedical and Health Informatics}, vol.~20, no.~1, pp. 304--321,
  2016.

\bibitem{szeliski1997creating}
R.~Szeliski and H.-Y. Shum, ``Creating full view panoramic image mosaics and
  environment maps,'' in \emph{Proceedings on Computer Graphics and Interactive
  Techniques}, 1997, pp. 251--258.

\bibitem{fischler1981random}
M.~A. Fischler and R.~C. Bolles, ``Random sample consensus: a paradigm for
  model fitting with applications to image analysis and automated
  cartography,'' \emph{Communications of the ACM}, vol.~24, no.~6, pp.
  381--395, 1981.

\bibitem{lowe1999object}
D.~G. Lowe, ``Object recognition from local scale-invariant features,'' in
  \emph{ICCV}, vol.~2.\hskip 1em plus 0.5em minus 0.4em\relax IEEE, 1999, pp.
  1150--1157.

\bibitem{brown2007automatic}
M.~Brown and D.~G. Lowe, ``Automatic panoramic image stitching using invariant
  features,'' \emph{International Journal of Computer Vision}, vol.~74, no.~1,
  pp. 59--73, 2007.

\bibitem{iakovidis2013efficient}
D.~K. Iakovidis, E.~Spyrou, and D.~Diamantis, ``Efficient homography-based
  video visualization for wireless capsule endoscopy,'' in \emph{IEEE
  International Conference on BioInformatics and BioEngineering}.\hskip 1em
  plus 0.5em minus 0.4em\relax IEEE, 2013, pp. 1--4.

\bibitem{bano2020deep}
S.~Bano, F.~Vasconcelos, M.~Tella-Amo, G.~Dwyer, C.~Gruijthuijsen,
  E.~Vander~Poorten, T.~Vercauteren, S.~Ourselin, J.~Deprest, and D.~Stoyanov,
  ``Deep learning-based fetoscopic mosaicking for field-of-view expansion,''
  \emph{International Journal of Computer Assisted Radiology and Surgery},
  vol.~15, no.~11, pp. 1807--1816, 2020.

\bibitem{zaragoza2013projective}
J.~Zaragoza, T.-J. Chin, M.~S. Brown, and D.~Suter, ``As-projective-as-possible
  image stitching with moving dlt,'' in \emph{CVPR}, 2013, pp. 2339--2346.

\bibitem{li2017quasi}
N.~Li, Y.~Xu, and C.~Wang, ``Quasi-homography warps in image stitching,''
  \emph{IEEE Transactions on Multimedia}, vol.~20, no.~6, pp. 1365--1375, 2017.

\bibitem{li2015dual}
S.~Li, L.~Yuan, J.~Sun, and L.~Quan, ``Dual-feature warping-based motion model
  estimation,'' in \emph{ICCV}, 2015, pp. 4283--4291.

\bibitem{zhang2014parallax}
F.~Zhang and F.~Liu, ``Parallax-tolerant image stitching,'' in \emph{CVPR},
  2014, pp. 3262--3269.

\bibitem{lee2020warping}
K.-Y. Lee and J.-Y. Sim, ``Warping residual based image stitching for large
  parallax,'' in \emph{CVPR}, 2020, pp. 8198--8206.

\bibitem{chen2016natural}
Y.-S. Chen and Y.-Y. Chuang, ``Natural image stitching with the global
  similarity prior,'' in \emph{ECCV}.\hskip 1em plus 0.5em minus 0.4em\relax
  Springer, 2016, pp. 186--201.

\bibitem{fan2019stereoscopic}
X.~Fan, J.~Lei, Y.~Fang, Q.~Huang, N.~Ling, and C.~Hou, ``Stereoscopic image
  stitching via disparity-constrained warping and blending,'' \emph{IEEE
  Transactions on Multimedia}, vol.~22, no.~3, pp. 655--665, 2019.

\bibitem{kose2017automated}
K.~Kose, M.~Gou, O.~Y{\'e}lamos, M.~Cordova, A.~M. Rossi, K.~S. Nehal
  \emph{et~al.}, ``Automated video-mosaicking approach for confocal microscopic
  imaging in vivo: an approach to address challenges in imaging living tissue
  and extend field of view,'' \emph{Scientific Reports}, vol.~7, no.~1, pp.
  1--11, 2017.

\bibitem{gong2020robust}
L.~Gong, J.~Zheng, Z.~Ping, Y.~Wang, S.~Wang, and S.~Zuo, ``Robust mosaicing of
  endomicroscopic videos via context-weighted correlation ratio,'' \emph{IEEE
  Transactions on Biomedical Engineering}, 2020.

\bibitem{vercauteren2006robust}
T.~Vercauteren, A.~Perchant, G.~Malandain, X.~Pennec, and N.~Ayache, ``Robust
  mosaicing with correction of motion distortions and tissue deformations for
  in vivo fibered microscopy,'' \emph{Medical Image Analysis}, vol.~10, no.~5,
  pp. 673--692, 2006.

\bibitem{bano2020deepMiccai}
S.~Bano, F.~Vasconcelos, L.~M. Shepherd, E.~Vander~Poorten, T.~Vercauteren,
  S.~Ourselin, A.~L. David, J.~Deprest, and D.~Stoyanov, ``Deep placental
  vessel segmentation for fetoscopic mosaicking,'' in \emph{MICCAI}.\hskip 1em
  plus 0.5em minus 0.4em\relax Springer, 2020, pp. 763--773.

\bibitem{gaisser2018stable}
F.~Gaisser, S.~H. Peeters, B.~A. Lenseigne, P.~P. Jonker, and D.~Oepkes,
  ``Stable image registration for in-vivo fetoscopic panorama reconstruction,''
  \emph{Journal of Imaging}, vol.~4, no.~1, p.~24, 2018.

\bibitem{newcombe2015dynamicfusion}
R.~A. Newcombe, D.~Fox, and S.~M. Seitz, ``Dynamicfusion: Reconstruction and
  tracking of non-rigid scenes in real-time,'' in \emph{CVPR}, 2015, pp.
  343--352.

\bibitem{totz2011dense}
J.~Totz, P.~Mountney, D.~Stoyanov, and G.-Z. Yang, ``Dense surface
  reconstruction for enhanced navigation in mis,'' in \emph{MICCAI}.\hskip 1em
  plus 0.5em minus 0.4em\relax Springer, 2011, pp. 89--96.

\bibitem{song2018mis}
J.~Song, J.~Wang, L.~Zhao, S.~Huang, and G.~Dissanayake, ``Mis-slam: Real-time
  large-scale dense deformable slam system in minimal invasive surgery based on
  heterogeneous computing,'' \emph{IEEE Robotics and Automation Letters},
  vol.~3, no.~4, pp. 4068--4075, 2018.

\bibitem{zhou2020smooth}
H.~Zhou and J.~Jayender, ``Smooth deformation field-based mismatch removal in
  real-time,'' \emph{arXiv preprint arXiv:2007.08553}, 2020.

\bibitem{soper2012surface}
T.~D. Soper, M.~P. Porter, and E.~J. Seibel, ``Surface mosaics of the bladder
  reconstructed from endoscopic video for automated surveillance,'' \emph{IEEE
  Transactions on Biomedical Engineering}, vol.~59, no.~6, pp. 1670--1680,
  2012.

\bibitem{richa2012hybrid}
R.~Richa, B.~V{\'a}gv{\"o}lgyi, M.~Balicki, G.~Hager, and R.~H. Taylor,
  ``Hybrid tracking and mosaicking for information augmentation in retinal
  surgery,'' in \emph{MICCAI}.\hskip 1em plus 0.5em minus 0.4em\relax Springer,
  2012, pp. 397--404.

\bibitem{karargyris2010three}
A.~Karargyris and N.~Bourbakis, ``Three-dimensional reconstruction of the
  digestive wall in capsule endoscopy videos using elastic video
  interpolation,'' \emph{IEEE Transactions on Medical Imaging}, vol.~30, no.~4,
  pp. 957--971, 2010.

\bibitem{zhou2019real}
H.~Zhou and J.~Jagadeesan, ``Real-time dense reconstruction of tissue surface
  from stereo optical video,'' \emph{IEEE Transactions on Medical Imaging},
  vol.~39, no.~2, pp. 400--412, 2019.

\bibitem{mahmoud2018live}
N.~Mahmoud, T.~Collins, A.~Hostettler, L.~Soler, C.~Doignon, and J.~M.~M.
  Montiel, ``Live tracking and dense reconstruction for handheld monocular
  endoscopy,'' \emph{IEEE Transactions on Medical Imaging}, vol.~38, no.~1, pp.
  79--89, 2018.

\bibitem{mountney2009dynamic}
P.~Mountney and G.-Z. Yang, ``Dynamic view expansion for minimally invasive
  surgery using simultaneous localization and mapping,'' in \emph{International
  Conference of the IEEE Engineering in Medicine and Biology Society}.\hskip
  1em plus 0.5em minus 0.4em\relax IEEE, 2009, pp. 1184--1187.

\bibitem{mountney2010motion}
P.~Mountney and Yang, ``Motion compensated slam for image guided surgery,'' in
  \emph{MICCAI}.\hskip 1em plus 0.5em minus 0.4em\relax Springer, 2010, pp.
  496--504.

\bibitem{lamarca2020defslam}
J.~Lamarca, S.~Parashar, A.~Bartoli, and J.~Montiel, ``Defslam: Tracking and
  mapping of deforming scenes from monocular sequences,'' \emph{IEEE
  Transactions on Robotics}, 2020.

\bibitem{poggi2019guided}
M.~Poggi, D.~Pallotti, F.~Tosi, and S.~Mattoccia, ``Guided stereo matching,''
  in \emph{CVPR}, 2019, pp. 979--988.

\bibitem{erden2012understanding}
M.~S. Erden, B.~Rosa, J.~Szewczyk, and G.~Morel, ``Understanding soft-tissue
  behavior for application to microlaparoscopic surface scan,'' \emph{IEEE
  Transactions on Biomedical Engineering}, vol.~60, no.~4, pp. 1059--1068,
  2012.

\bibitem{hughes2015high}
M.~Hughes and G.-Z. Yang, ``High speed, line-scanning, fiber bundle
  fluorescence confocal endomicroscopy for improved mosaicking,''
  \emph{Biomedical Optics Express}, vol.~6, no.~4, pp. 1241--1252, 2015.

\bibitem{rosa2012building}
B.~Rosa, M.~S. Erden, T.~Vercauteren, B.~Herman, J.~Szewczyk, and G.~Morel,
  ``Building large mosaics of confocal edomicroscopic images using visual
  servoing,'' \emph{IEEE Transactions on Biomedical Engineering}, vol.~60,
  no.~4, pp. 1041--1049, 2012.

\bibitem{loewke2010vivo}
K.~E. Loewke, D.~B. Camarillo, W.~Piyawattanametha, M.~J. Mandella, C.~H.
  Contag, S.~Thrun, and J.~K. Salisbury, ``In vivo micro-image mosaicing,''
  \emph{IEEE Transactions on Biomedical Engineering}, vol.~58, no.~1, pp.
  159--171, 2010.

\bibitem{perperidis2020image}
A.~Perperidis, K.~Dhaliwal, S.~McLaughlin, and T.~Vercauteren, ``Image
  computing for fibre-bundle endomicroscopy: A review,'' \emph{Medical Image
  Analysis}, vol.~62, p. 101620, 2020.

\bibitem{innmann2016volumedeform}
M.~Innmann, M.~Zollh{\"o}fer, M.~Nie{\ss}ner, C.~Theobalt, and M.~Stamminger,
  ``Volumedeform: Real-time volumetric non-rigid reconstruction,'' in
  \emph{ECCV}.\hskip 1em plus 0.5em minus 0.4em\relax Springer, 2016, pp.
  362--379.

\bibitem{kavan2008geometric}
L.~Kavan, S.~Collins, J.~{\v{Z}}{\'a}ra, and C.~O'Sullivan, ``Geometric
  skinning with approximate dual quaternion blending,'' \emph{ACM Transactions
  on Graphics}, vol.~27, no.~4, pp. 1--23, 2008.

\bibitem{rublee2011orb}
E.~Rublee, V.~Rabaud, K.~Konolige, and G.~Bradski, ``Orb: An efficient
  alternative to sift or surf,'' in \emph{ICCV}.\hskip 1em plus 0.5em minus
  0.4em\relax IEEE, 2011, pp. 2564--2571.

\bibitem{bay2006surf}
H.~Bay, T.~Tuytelaars, and L.~Van~Gool, ``Surf: Speeded up robust features,''
  in \emph{ECCV}.\hskip 1em plus 0.5em minus 0.4em\relax Springer, 2006, pp.
  404--417.

\bibitem{mur2015orb}
R.~Mur-Artal, J.~M.~M. Montiel, and J.~D. Tardos, ``Orb-slam: a versatile and
  accurate monocular slam system,'' \emph{IEEE Transactions on Robotics},
  vol.~31, no.~5, pp. 1147--1163, 2015.

\bibitem{schonberger2017comparative}
J.~L. Schonberger, H.~Hardmeier, T.~Sattler, and M.~Pollefeys, ``Comparative
  evaluation of hand-crafted and learned local features,'' in \emph{CVPR},
  2017, pp. 1482--1491.

\bibitem{sun2004multi}
S.-L. Sun and Z.-L. Deng, ``Multi-sensor optimal information fusion kalman
  filter,'' \emph{Automatica}, vol.~40, no.~6, pp. 1017--1023, 2004.

\bibitem{sorkine2007rigid}
O.~Sorkine and M.~Alexa, ``As-rigid-as-possible surface modeling,'' in
  \emph{Symposium on Geometry processing}, vol.~4, 2007, pp. 109--116.

\bibitem{arun1987least}
K.~S. Arun, T.~S. Huang, and S.~D. Blostein, ``Least-squares fitting of two 3-d
  point sets,'' \emph{IEEE Transactions on Pattern Analysis and Machine
  Intelligence}, no.~5, pp. 698--700, 1987.

\bibitem{curless1996volumetric}
B.~Curless and M.~Levoy, ``A volumetric method for building complex models from
  range images,'' in \emph{Annual conference on Computer Graphics and
  Interactive Techniques}, 1996, pp. 303--312.

\bibitem{lin2015adaptive}
C.-C. Lin, S.~U. Pankanti, K.~Natesan~Ramamurthy, and A.~Y. Aravkin, ``Adaptive
  as-natural-as-possible image stitching,'' in \emph{CVPR}, 2015, pp.
  1155--1163.

\end{thebibliography}

\end{document}